\documentclass[conference]{IEEEtran}
\IEEEoverridecommandlockouts
\usepackage{iftex}
\ifXeTeX\usepackage[OT1]{fontenc}\fi

\usepackage{cite}
\usepackage{amsmath,amssymb,amsfonts}
\usepackage{graphicx}
\usepackage{booktabs}
\usepackage{textcomp}
\usepackage{booktabs,multirow}
\usepackage[table]{xcolor}
\usepackage{url}
\usepackage{tabularx}
\usepackage{makecell}
\usepackage{siunitx}
\begin{document}
\bstctlcite{IEEEexample:BSTcontrol} % references: 6+ authors -> "First Author et al."

\title{MetricPose: Flexible 3D Human Pose Estimation with Depth and Geometric Cues}

\author{
\IEEEauthorblockN{
Shifan Zhu, Sabad Modi, Hang Yang, Arushi Agrawal, Steven Gao,\\
Meghan Huber, Nathan Wycoff, and Donghyun Kim$^{\dagger}$
}
\IEEEauthorblockA{
University of Massachusetts Amherst, Amherst, MA, USA
}
\thanks{$^{\dagger}$Corresponding author: Donghyun Kim, \texttt{donghyunkim@umass.edu}.}
}

\maketitle

\begin{abstract}

In this paper\footnote{Project website: \url{https://metric3db.github.io/metric-3db}}, we present a 3D human pose estimator that predicts metrically accurate human joint positions in the camera coordinate frame. This task requires recovering the configuration of the body joints, scale, and metric camera-local placement. In monocular images, body size and camera distance are coupled through perspective projection and therefore cannot be determined uniquely from image alone. This ambiguity can be reduced by combining three complementary sources of information: learned RGB priors, direct depth measurements, and geometric constraints (e.g., ground contact) from the environment. We extend a pretrained full-body estimator with optional metric depth- and geometry-processing modules that improve camera-distance estimation and depth-guided body refinement while preserving RGB-only operation. The resulting unified model supports RGB, RGB-Ground, and RGB-D using shared learned parameters. Unlike pelvis-relative error, camera-local error also measures placement relative to the camera. Compared with the existing foundation model SAM 3D Body, our methods with RGB, RGB-Ground, and RGB-D modes reduce camera-local joint-position error by \textbf{12.16\%}, \textbf{31.00\%} and \textbf{54.61\%} respectively.

\end{abstract}


\section{Introduction}

Recent monocular 3D human pose estimators can recover body-joint configurations accurately under occlusion, truncation, and unusual poses~\cite{sam3dbody,fast3dbody}. Most progress, however, is measured after subtracting the pelvis and, in some cases, aligning body scale. These protocols isolate root-relative pose but discard errors in lateral position, camera distance, and physical scale. Camera-local evaluation instead compares predicted and ground-truth joints directly in camera coordinates, without translation, rotation, or scale alignment~\cite{metrabs}. This distinction matters when applications depend on metric distances and motion. Learning collaborative behavior from human demonstrations, such as two people carrying a table, requires preserving their relative positions and trajectories in a common metric frame~\cite{ng2023cooperative}. Social-interaction analysis similarly depends on interpersonal distances and spatial arrangements~\cite{setti2015fformation}, while biomechanical measures require physical scale and global displacement~\cite{opencap}. Pelvis centering and scale alignment remove these quantities~\cite{metrabs}.

\begin{figure}[!t]
\centering
\includegraphics[width=\columnwidth]{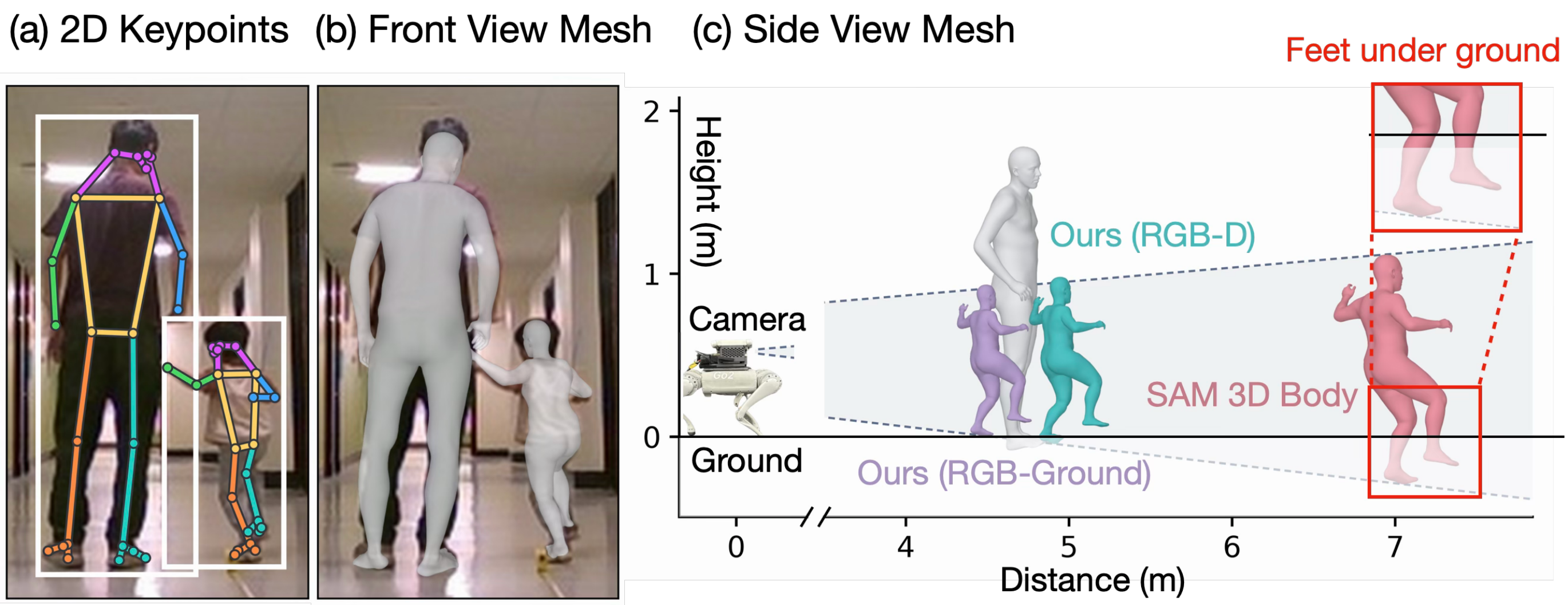}
\caption{Monocular scale--distance ambiguity. (a) Input image and 2D keypoints. (b) Different body scales and distances yield similar projections. (c) Camera-coordinate visualization exposes errors in physical scale, distance, and ground contact, which metric depth and ground geometry help resolve.}
\vspace{-10pt}
\label{fig:scale-ambiguity}
\end{figure}

Metric 3D human pose estimation in camera coordinates requires recovering body-joint configuration, physical scale, and camera-relative placement, with scale and placement coupled under perspective projection because body size and camera distance jointly determine the projected image extent~\cite{rootnet}. Monocular estimators commonly address this ambiguity using priors learned from large-scale in-the-wild image collections. These priors encode typical body proportions, poses, and camera configurations, but their reliability depends on the training distribution. For underrepresented body shapes, such as children, a model may explain the image as an adult-sized body farther from the camera, producing a plausible image-space reconstruction with incorrect physical scale and placement, as illustrated in Fig.~\ref{fig:scale-ambiguity}.

Measured depth provides direct distance information, helping constrain camera distance and resolve the size--distance ambiguity. Depth is also not the only source of metric information available at inference. In many fixed or robot-mounted systems, camera height is known from the mounting configuration, while gravity is available from calibration or an onboard inertial sensor. Together, they define the ground plane in camera coordinates and, under a valid foot--ground contact, constrain body scale and camera-relative placement.

Our key insight is therefore to use RGB priors, measured depth, and ground geometry at complementary stages of reconstruction. Building on SAM 3D Body (3DB)~\cite{sam3dbody}, learned RGB-D feature fusion incorporates depth context, a two-pass local depth-residual module refines pose from discrepancies between observed surface depth and predicted landmark depth, and geometric corrections adjust metric scale and camera-relative placement. These components share one parameter set and activate according to input availability, supporting RGB, RGB-Ground and RGB-D.

We evaluate the unified model across all three input modes. We report mean per-joint position error (MPJPE), the average Euclidean distance between corresponding predicted and ground-truth 3D joints, under three protocols. Pelvis-relative MPJPE centers both skeletons at the pelvis, removing global translation; Procrustes-aligned MPJPE (PA-MPJPE) additionally removes global rotation and isotropic scale; and camera-local MPJPE applies no alignment, so it also penalizes errors in camera distance and lateral position, including those caused by incorrect body scale. The main contributions are:
\begin{itemize}
\item \textbf{Unified optional-input estimator.} One shared-weight 3DB-based model operates with RGB, RGB-Ground, or RGB-D by activating the available learned and geometric pathways.
\item \textbf{Complementary depth and geometry processing.} Learned depth fusion and a two-pass local-residual refiner improve pose, while deterministic depth-surface or ground-contact corrections jointly rescale the body and camera translation while preserving the 2D projection.
\item \textbf{Improved camera-local accuracy.} Our RGB, RGB-Ground, and RGB-D modes reduce errors in physical scale and camera-relative placement that aligned metrics obscure, lowering camera-local MPJPE by 12.2\%, 31.0\%, and 54.6\%, respectively, relative to released 3DB\footnote{Geometric mean of camera-local error ratios across seven evaluations (Tables \ref{tab:standard}--\ref{tab:rgbd}, including three with MoGe-2) for RGB and RGB-D, and four evaluations (Table \ref{tab:rgbd}) for RGB-Ground.}.
In a controlled simulation with a small-stature avatar \SI{5}{\meter} from the camera, SAM 3D Body misestimates its distance by more than \SI{4}{\meter}, whereas our RGB-D mode reduces this error to about \SI{7}{\centi\meter}.
\end{itemize}

\section{Related Work}

Monocular human reconstruction estimates body pose and shape from a single RGB image, but monocular evidence alone cannot separate physical body scale from camera distance. Prior work therefore relies on learned priors or adds depth, scene geometry, or optional sensing.

\textbf{Monocular human reconstruction and camera-local placement.} Methods such as MHFormer and HybrIK recover body-joint configuration, which is commonly evaluated relative to the pelvis~\cite{mhformer,hybrik}. RootNet estimates root depth from an assumed physical person size~\cite{rootnet}, while MeTRAbs and NLF combine 2D and metric 3D predictions with camera intrinsics to obtain absolute poses~\cite{metrabs,nlf}. SPEC and CameraHMR estimate camera parameters~\cite{spec,camerahmr}, which corrects perspective but not the size--distance ambiguity, and BLADE regresses the person's pelvis depth and then solves for focal length and translation at close range~\cite{blade}. Multi-HMR predicts multiple whole-body meshes and their camera-local locations~\cite{multihmr}, while Metric-Aware HMR builds metric pseudo-ground truth by jointly optimizing crowd translations with height priors and monocular depth~\cite{metricawarehmr}. More recently, 3DB combines large-scale visual pretraining, promptable decoding, and a structured human representation~\cite{sam3dbody}, and Fast 3DB improves its computational efficiency~\cite{fast3dbody}. For all-age scenes, BEV estimates age groups to order people in depth~\cite{bev}, and Anny-Fit fits an all-age body model in camera space using VLM-estimated age and predicted metric depth~\cite{annyfit}. These methods infer metric scale from image evidence, average body-size priors, or predicted depth. Our method corrects the inherited RGB estimate with measured depth or known ground geometry, processing each detected person independently.

\textbf{Depth-assisted human pose and mesh recovery.} Depth sensing provides direct metric observations of the visible body surface. Existing methods integrate RGB and depth features volumetrically~\cite{zimmermann}, reconstruct bodies from back-projected partial point clouds~\cite{m3}, lift joints detected in RGB using depth sampled at their image locations~\cite{simpledepthpose}, or fit parametric bodies to image keypoints and measured point clouds~\cite{fang}. RGB-D sensing has also been used for unified body, hand, and face estimation~\cite{bashirov}. RELI11D shows that pelvis-relative accuracy does not guarantee global accuracy~\cite{reli11d}. Nevertheless, measured depth represents the visible clothing or skin surface rather than the underlying anatomical joint centers. Existing RGB-D approaches also generally require depth during every inference. In contrast, our model learns to use depth without sacrificing overall RGB-only accuracy.

\textbf{Scene geometry and physical constraints.} Scene information and physical contact provide constraints on human placement. Classical single-view metrology recovers physical measurements and camera location from reference-plane and vanishing-point geometry~\cite{criminisi1999metrology}. Fei et al. estimate camera intrinsics, the ground plane, and distances between people from their poses, assuming an average human height~\cite{fei2021distance}. SA-HMR uses human--scene contact to recover absolute body position in a pre-scanned scene observed by a localized camera~\cite{sahmr}. IPMAN introduces contact and pressure consistency with a ground plane to improve physical plausibility~\cite{ipman}. Our ground correction instead takes the plane from known camera height and gravity, needing no scene scan or height prior, and uses foot contact to rescale body size and camera distance together.

\textbf{Human reconstruction with optional sensing.} Depth has been used as privileged information during training before being removed at inference~\cite{simoni}, although such cross-modal transfer is not always beneficial~\cite{xue2023mfh}. Modality dropout trains networks to tolerate missing inputs~\cite{moddrop}, and we adopt it for depth. PromptHMR accepts optional spatial and semantic prompts and camera intrinsics~\cite{prompthmr}, but unlike depth and ground geometry, none of these inputs constrains metric scale or distance. Privileged-information methods cannot use depth when it becomes available during deployment, while existing RGB-D pose estimators assume fixed inputs~\cite{zimmermann,bashirov}.% Our unified model instead uses depth and ground information when provided while preserving operation without either input.

Overall, previous methods generally rely on monocular priors, require metric sensing, or need scene scans or height priors to exploit scene geometry. Our method combines these sources for metric camera-local reconstruction while preserving operation when optional metric inputs are unavailable.

\begin{figure*}[!t]
\centering
% Previous architecture figure retained as figs/architecture.png.
\includegraphics[width=\textwidth]{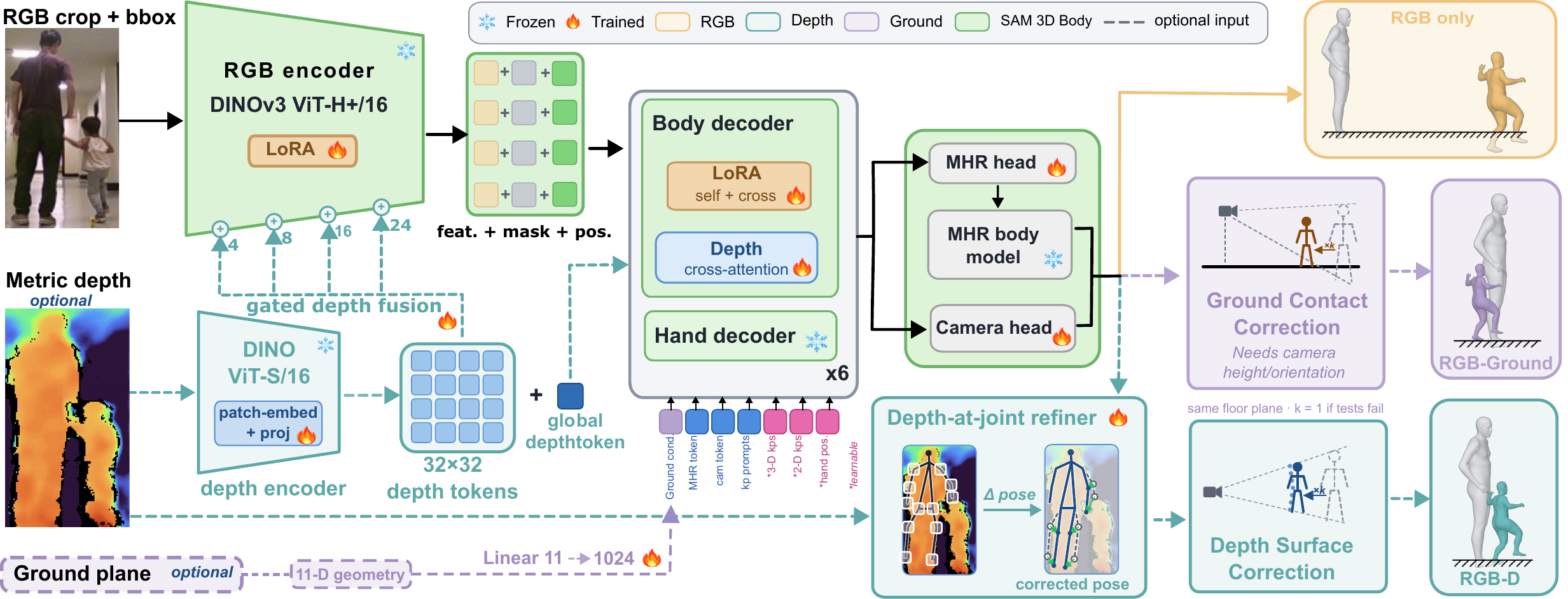}
\vspace{-10pt}
\caption{Unified architecture for RGB, RGB-Ground, and RGB-D with
shared weights. SAM 3D Body provides the body and camera estimate;
optional ground geometry applies a ground-contact correction, while optional
depth adds feature fusion, joint-level pose refinement, and depth-surface
metric correction.}
\vspace{-10pt}
\label{fig:architecture}
\end{figure*}

\section{Method}
\label{sec:method}

We extend 3DB with optional metric depth and ground geometry while retaining RGB-only operation. Our objective is to recover each detected person's metrically scaled 3D body and place it correctly relative to the camera. For person $n$, the predicted camera-frame position of joint $j$ is $\hat{\mathbf d}_{n,j}^{\mathrm{cam}}=\hat{\mathbf d}_{n,j}^{\mathrm{pel}}+\hat{\mathbf t}_n$, where $\hat{\mathbf d}_{n,j}^{\mathrm{pel}}$ is the metrically scaled pelvis-relative prediction and $\hat{\mathbf t}_n$ is the camera-frame pelvis position.

\subsection{3DB Backbone and Adaptation}
\label{sec:pretrained}

We build on 3DB~\cite{sam3dbody}. Given a $512\times512$ person crop, its DINOv3 ViT-H/16+ encoder produces image tokens processed by a six-layer promptable body decoder. A trainable MLP maps the decoded pose token to Momentum Human Rig (MHR) parameters, including global orientation, body pose, shape, and skeletal scale, while the frozen, differentiable MHR body model converts them into a mesh and MHR70 joints. A separate perspective-camera head predicts crop-relative camera parameters, which are converted to camera-local translation using the person box and effective intrinsics.

We apply LoRA to the pretrained image encoder and body decoder, while training the MHR and camera heads and the new depth, ground, and pose-refinement modules. In the image encoder, we use rank-16 LoRA for attention and rank-8 LoRA for feed-forward layers; in the body decoder, we use rank-8 LoRA for self- and cross-attention. The remaining pretrained encoder and decoder weights, the hand branch, and the MHR body model remain frozen. The camera head receives a stop-gradient pose representation so that placement losses do not update the body representation through this branch.

\subsection{Depth Encoding and Cross-Modal Feature Integration}
\label{sec:depth}

Depth enters the model through complementary feature-level and measurement-level pathways. The feature-level pathway encodes depth into spatial and global tokens that condition the RGB representation through encoder residual writes and decoder cross-attention. The measurement-level pathways use metric depth: local residuals refine body pose in Section~\ref{sec:refinement}, while surface agreement corrects metric scale and camera-local placement in Section~\ref{sec:anchors}. To construct the feature-level pathway, we align depth with the RGB crop and encode bounded depth, camera rays, and validity using a small pretrained trunk. Depths outside $[D_{\min},D_{\max}]$ are marked invalid.

For crop pixel $p=(u,v)$ of person $n$, let $D_{n,p}$ denote depth in meters and $V_{n,p}\in\{0,1\}$ denote its validity. Using cropped intrinsics $K_{c,n}$, we inverse-project each pixel to bounded normalized camera-ray coordinates $\boldsymbol\rho_{n,p}\in[-1,1]^2$.

Following~\cite{vanishingdepth}, we encode metric depth using the global Sinusoidal Depth Processing (SDP) representation. We augment it with the projective-ray coordinates and validity:
\begin{equation*}
\mathbf g_{n,p}
=
\left[
\operatorname{SDP}(D_{n,p}),
\rho_{n,p,x}V_{n,p},
\rho_{n,p,y}V_{n,p},
V_{n,p}
\right]
\in\mathbb R^{C+3}.
\end{equation*}

We use $C=32$ SDP channels with temperature $3\times10^{-4}$ and $D_{\max}=\SI{15}{\meter}$. We set $D_{\min}=\SI{0.25}{\meter}$ for validity and use $r_{\max}=2$ for ray clipping. SDP is computed from absolute depth without validity masking, while $V_{n,p}$ explicitly identifies invalid pixels. The SDP formulation follows~\cite{vanishingdepth}.

A ViT-S/16 (12 layers, 384 channels) initialized from RGB pretraining maps $\mathbf g_n$ to a $32\times32$ grid of tokens, linearly projected to a 1280-dim RGB token to give depth tokens $\mathbf z_n$ on the RGB feature grid without resampling. Its transformer trunk is frozen while its patch embedding and depth-specific projections are trained. A linearly projected validity-weighted mean forms a global depth token, and an auxiliary decoder reconstructs the normalized depth target under $\mathcal L_{\mathrm{rec}}$.

At encoder blocks $\ell\in\{4,8,16,24\}$, let $\mathbf x_{n,\ell}$ denote the RGB tokens entering block $\ell$. The projected depth write is matched to the local RGB-token magnitude and gated by token reliability and a channel-wise gate:
\begin{equation}
\begin{gathered}
\widetilde{\mathbf z}_{n,\ell}
=
\frac{\lVert\mathbf x_{n,\ell}\rVert_2}
{\max(\lVert P_\ell\mathbf z_n\rVert_2,\epsilon)}
P_\ell\mathbf z_n,\\
\Delta\mathbf x_{n,\ell}
=
\boldsymbol\eta_\ell\odot
\left(a_{n,\ell}\widetilde{\mathbf z}_{n,\ell}\right).
\end{gathered}
\label{eq:fusion-residual}
\end{equation}

Here, $P_\ell$ is a learned linear projection, and $a_{n,\ell}\in(0,1)$ is a scalar reliability weight per token predicted from the normalized RGB and depth tokens and their valid-depth fraction $\bar{\mathbf v}_n$. The channel-wise gate is $\boldsymbol\eta_\ell=(1-c)\tanh(\boldsymbol\gamma_\ell)+c$, with $\boldsymbol\gamma_\ell$ initialized to $0.15$ and $c=0.05$. Norms are computed per token, with $\epsilon=10^{-6}$ preventing division by zero; norm matching keeps the relative write magnitude comparable across backbone depths.

We add the depth correction residually as $\mathbf x'_{n,\ell}=\mathbf x_{n,\ell}+m_{\mathrm{enc},n}\Delta\mathbf x_{n,\ell}$, where $m_{\mathrm{enc},n}$ is one when usable depth is available and zero otherwise. Each body-decoder layer additionally attends to the spatial and global depth tokens. Spatial tokens without valid depth are masked, and sample-level availability gates the complete depth write. Consequently, both backbone and decoder depth writes vanish without usable depth, recovering the adapted RGB path.

\subsection{Depth-Guided Pose Refinement}
\label{sec:refinement}

Global RGB-D fusion provides distributed depth context but does not explicitly compare measured and predicted depth around individual body keypoints. We therefore use a two-pass depth-guided module to refine pose before the metric-placement corrections of Section~\ref{sec:anchors}. The first MHR pass predicts body-keypoint positions $\{\hat{\mathbf d}_{n,j}^{(1)}\}_j$, a mesh, and a camera translation $\hat{\mathbf t}_n^{(1)}$. We project selected body keypoints into the registered depth crop and sample a local residual grid around each projection. For selected keypoint $j$, we form the local measured-minus-predicted camera-depth residual $R_{n,j}(u,v)=D_n(u,v)-\left(\hat d_{n,j,z}^{(1)}+\hat t_{z,n}^{(1)}\right)$. For each person, we subtract the median of the valid center-pixel residuals across selected keypoints and robustly normalize the remaining residuals. Removing this shared component preserves global translation for the metric-placement stage and treats local deviations as evidence for pose refinement.

For each selected body keypoint, the refiner encodes the local depth-residual and validity patches together with the sampled RGB feature, hip-relative keypoint depth, image location, and keypoint identity into a transformer input vector. A separate global input vector combines the first-pass pose representation with the per-person median residual and valid-patch fraction. A lightweight transformer jointly processes these vectors and predicts a zero-initialized, depth-gated update, restricted to global orientation and body pose; shape, scale, hand, and face parameters are held fixed. Therefore, the local depth-guided refinement cannot change metric body size, which is handled by the surface-matching stage in Section~\ref{sec:anchors}. The first-pass camera parameters are reused, and camera-local translation is recomputed for the refined body. Refinement is applied once, at the final decoder layer. Missing depth zeros the update, while invalid local samples are masked.

We run the MHR head a second time using the refined pose initialization, after which the frozen MHR body model regenerates the mesh and MHR70 joints. The ground correction and, at inference, the depth-surface correction of Section~\ref{sec:anchors} are applied after this second pass. All residual statistics and updates are computed independently for each person.

\subsection{Metric Scale and Camera-Local Placement}
\label{sec:anchors}

The scale factors in this subsection are computed deterministically from the learned body and camera predictions and the available scene measurements; no additional network directly regresses these corrections.

The inherited camera head predicts weak-perspective scale $s_n$ and crop-relative offsets for person $n$. Let $f_n$ and $b_n$ denote the effective focal length and person-box size, respectively. Let $\hat L_n$ be the sum of eight SMPL bone lengths regressed from the predicted mesh, and let $L_{\mathrm{ref}}=\SI{2.441}{\meter}$ be a calibration constant selected from the starting checkpoint. We define $\kappa_{\mathrm{rgb},n}=\operatorname{clip}(\hat L_n/L_{\mathrm{ref}},0.25,4.0)$, which is approximately one for a typical starting-checkpoint prediction. The constant $L_{\mathrm{ref}}$ is neither the person's height nor an assumed population-average body size; it only fixes the scale convention between body size and camera translation. The RGB placement is
\begin{equation*}
\hat t_{z,n}^{\mathrm{cliff}}
=
\operatorname{clip}\!\left(
\frac{2f_n}{b_ns_n},t_{\min},t_{\max}
\right),
\qquad
\hat{\mathbf t}_n^{\mathrm{rgb}}
=
\kappa_{\mathrm{rgb},n}\hat{\mathbf t}_n^{\mathrm{cliff}}.
\end{equation*}
The lateral components use the predicted offsets and crop-center displacement. Intrinsics are supplied when available and otherwise approximated from an estimated or default field of view. A proportionally larger predicted body is therefore placed proportionally farther from the camera, preserving its projected image size.

Learned depth conditioning estimates pose and coarse placement. At inference, the deterministic surface correction then uses registered depth to apply a residual global similarity scale without changing pose. We construct a point-splat depth rendering $\hat D_n(p)$ from the predicted vertices and compare it with valid measured depth in an eroded, depth-consistent support region $\Omega_n$. The robust depth-surface scale is
\begin{equation}
\log\kappa_{D,n}
=
\operatorname*{median}_{p\in\Omega_n}
\left[
\log D_n(p)-\log\hat D_n(p)
\right].
\label{eq:surface-similarity}
\end{equation}

When accepted, $\kappa_{D,n}$ multiplies the pelvis-relative joints, mesh vertices, and camera translation. It therefore changes metric scale and placement without changing the body-joint configuration or the 2D projection.

A ground cue is represented by the camera-local plane $\mathbf n_g^\top\mathbf X+d=0$. A learned projection of floor-plane and ray-intersection features is added to the pose token. For geometric correction, $\mathbf P_{G,n}$ is the mean of the ten lowest mapped SMPL foot vertices under the current camera placement. The scale required to place this contact on the floor is $\kappa_{G,n}=d/(-\mathbf n_g^\top\mathbf P_{G,n})$.

The ground correction is applied only in RGB-Ground mode when the floor and contact geometry pass the validity tests; it does not require a learned contact classifier. The depth-surface correction is applied only in RGB-D mode when its validity tests pass. Each accepted correction multiplies the body and camera translation by the same factor. Ground and depth corrections are never applied together, and RGB mode uses neither correction. A rejected correction leaves the prediction unchanged. All three modes share one parameter set.

\subsection{Training Objective}
\label{sec:training}

Training uses source-homogeneous BEDLAM2~\cite{tesch2025bedlam2} and CMU Panoptic~\cite{Joo_2017_TPAMIpanoptic} batches, with each loss applied only where its required annotations are available. For MHR pass $k\in\{1,2\}$, we group the task objective as
\begin{equation}
\mathcal L_{\mathrm{task}}^{(k)}
=
\mathcal L_{\mathrm{place}}^{(k)}
+\mathbf 1_B\mathcal L_B^{(k)}
+\mathbf 1_P\mathcal L_P^{(k)}
+0.05\mathcal L_{\mathrm{ori}}^{(k)}
+\mathcal L_{\mathrm{pres}}^{(k)}.
\label{eq:task-loss}
\end{equation}

Here $\mathbf 1_B$ and $\mathbf 1_P$ select BEDLAM2 and Panoptic. The placement term supervises normalized lateral root position and logarithmic root depth. $\mathcal L_B$ combines mesh-regressed joint, vertex, reprojection, along-ray, and body-size supervision, whereas $\mathcal L_P$ combines 2D reprojection and size, neck, and along-ray supervision. The preservation term uses the frozen released model to constrain outputs without corresponding annotations. Each component is normalized over its valid annotations. The complete objective is
\begin{equation}
\begin{aligned}
\mathcal L
={}&
\mathcal L_{\mathrm{task}}^{(2)}
+\mathcal L_{\mathrm{frame}}^{(2)}
+\mathcal L_{\mathrm{own}}^{(2)}\\
&+0.5\left(
\mathcal L_{\mathrm{task}}^{(1)}
+\mathcal L_{\mathrm{frame}}^{(1)}
\right)
+0.25\mathcal L_{\mathrm{rec}}.
\end{aligned}
\label{eq:objective}
\end{equation}

The reconstruction term supervises the auxiliary depth decoder. For each pass, the framing term enforces agreement in projected size, metric body size, and root-relative pose across two crops of the same person. The second-pass own-placement term supervises the pre-correction camera root against ground truth on samples for which the ground correction is accepted.

\section{Experiments}
\label{sec:exp}

\subsection{Experimental Setup}
\label{sec:protocol}
\textbf{Datasets.} We fine-tune 3DB on equally sampled, source-homogeneous batches from BEDLAM2 and CMU Panoptic, both providing metric depth registered to RGB. BEDLAM2 depth is rendered in the RGB frame, while Panoptic sensor depth is projected using the provided calibration and temporal associations. Invalid projections are masked. We evaluate 3DPW~\cite{vonMarcard20183dpw}, EMDB~\cite{kaufmann2023emdb}, and RICH~\cite{Huang:CVPR:2022rich}, with depth-enabled results on held-out BEDLAM2, CMU Panoptic, HuMMan~\cite{cai2022humman}, and EgoBody~\cite{zhang2022egobody}.

\textbf{Geometric-correction settings.} The depth-surface similarity correction uses a two-pixel point-splat radius, a three-pixel-eroded support mask, the ratio band $0.5<D/\hat D<2$, at least 50 support pixels, and a maximum median absolute log residual of $0.1$. The ground correction requires the contact direction to lie more than $5^\circ$ below the floor horizon and accepts factors in $(0.5,2)$.

\textbf{Pelvis-distance correction settings.} The optional depth-based pelvis-distance correction adjusts camera-axis translation, and therefore pelvis depth, using visible-surface depth residuals. It uses a \SI{0.6}{\meter} band around the predicted pelvis depth, requires at least 64 visible mesh vertices, and abstains when the median absolute residual exceeds \SI{0.1}{\meter}.

\textbf{Joint-refinement settings.} The refiner uses $24\times24$ residual grids sampled at two-pixel intervals, clips centered residuals to $\pm$\SI{0.5}{\meter} and scales them by \SI{0.25}{\meter}, and employs a three-layer, four-head transformer.

\textbf{Training augmentation.} We use crop-scale jitter in $[0.8,1.3]$, $5\%$ translation, $\pm15^\circ$ in-plane rotation, RGB photometric augmentation, sensor-specific depth corruption, and up to two pixels of RGB-depth misregistration. Horizontal reflection is disabled, and full depth-modality dropout is linearly increased to $15\%$ during the first $30\%$ of training. Without depth, depth-dependent updates are gated off; ground processing remains active when ground geometry is provided.

\textbf{Metrics.} We report pelvis-relative MPJPE, PA-MPJPE, and camera-local MPJPE in millimeters, using pelvis centering, Procrustes alignment, and no alignment, respectively. Camera-local MPJPE retains pose, scale, and placement errors.

\textbf{Comparison protocol.} All reevaluated methods use identical oracle boxes and crops. We compare released 3DB, our shared-weight input modes, the optional training-free pelvis-distance correction, and published Fast 3DB results where available. RGB and RGB-D differ only in depth availability. RGB-D results use paired samples, with uncertainty estimated by bootstrapping acquisition or sequence groups.

\newcommand{\change}[2]{%
  \if#1+%
    \;{\color{red!70!black}(\ensuremath{+}#2\%)}%
  \else
    \;{\color{green!60!black}(\ensuremath{-}#2\%)}%
  \fi
}

\begin{table*}[!t]
  \caption{Results on 3DPW, EMDB, and RICH: MPJPE / PA-MPJPE / camera-local MPJPE (mm; $\downarrow$ lower is better).}
  \label{tab:standard}
  \centering
  % \resizebox{0.9\textwidth}{!}{%
  
  \setlength{\tabcolsep}{11pt}
  \begin{tabularx}{0.9\textwidth}{@{}lcccc@{}}
  % \begin{tabular}{@{}lcccc@{}}
  \toprule
  \multicolumn{1}{c}{\makecell[cc]{Method}} & \multicolumn{1}{c}{\makecell[cc]{Intrinsics}} & \makecell[cc]{3DPW (14)\\MPJPE / PA / Cam.} & \makecell[cc]{EMDB (24)\\MPJPE / PA / Cam.} & \makecell[cc]{RICH (24)\\MPJPE / PA / Cam.} \\
  \midrule
  CameraHMR & -- & 56.0 / 35.1 / -- & 70.3 / 43.3 / -- & 55.7 / 48.1 / -- \\
  PromptHMR & -- & 58.7 / 36.1 / -- & 71.7 / 41.0 / -- & 56.6 / 37.3 / -- \\
  NLF-L+fit$^{*}$ & -- & 54.9 / 33.6 / -- & 68.4 / 40.9 / -- & 51.0 / \textbf{28.7} / --\\
  Fast 3DB & -- & 59.5 / \textbf{30.4} / -- & 64.3 / 37.3 / -- & 58.8 / 33.8 / -- \\
  \midrule
  \multirow{2}{*}{SAM 3D Body}
   & GT & 55.0 / 33.7 / 248.4 & 63.1 / 38.8 / 123.4 & 52.8 / 31.8 / 342.2 \\
   & MoGe-2 & 55.1 / 33.7 / 511.2 & 63.1 / 38.8 / 139.4 & 54.9 / 32.2 / 607.7 \\
  \midrule
  \multirow{2}{*}{Ours (RGB)}
   & GT & 55.5 / 36.4 / 195.2 \change{-}{21.4} & 63.0 / 37.7 / 112.2 \change{-}{9.1} & 56.5 / 30.0 / 436.9 \change{+}{27.7} \\
   & MoGe-2 & 55.7 / 36.4 / 438.3 \change{-}{14.3} & 63.0 / 37.7 / 141.1 \change{+}{1.2} & 58.1 / 30.3 / 480.5 \change{-}{20.9} \\
  \multirow{2}{*}{Ours (RGB-D)}
   & GT & \textbf{50.0} / 34.4 / \textbf{103.4} \change{-}{58.4} & \textbf{53.3} / \textbf{35.7} / \textbf{49.9} \change{-}{59.6} & \textbf{41.9} / 29.4 / \textbf{50.4} \change{-}{85.3} \\
   & MoGe-2 & 51.7 / 35.4 / 129.3 \change{-}{74.7} & 58.7 / 35.8 / 60.2 \change{-}{56.8} & 57.0 / 30.5 / 175.3 \change{-}{71.2} \\
  \midrule

  \end{tabularx}% 
  % }
  % \par\vspace{2pt}
  \par\vspace{5pt}
  \begin{minipage}{0.9\textwidth}
  \footnotesize 
  \textit{Notes.} GT and MoGe-2~\cite{moge2} denote dataset-provided and estimated intrinsics, respectively. The percentage reports the change in camera-local MPJPE when switching from 3DB to our method. \textbf{Bold} marks the best result per metric. Published baselines are from~\cite{sam3dbody,fast3dbody}, and our reevaluated rows use oracle person boxes. $^{*}$ Trained on RICH. $^{\dagger}$ Public checkpoint. -- Value not reported.
  \end{minipage}
\vspace{-10pt}
\end{table*}

\begin{table*}[!t]
    \caption{Results on depth-enabled benchmarks: MPJPE / PA-MPJPE / camera-local MPJPE (mm; $\downarrow$). Best per metric in \textbf{bold}.}
    \label{tab:rgbd} 
    \centering 
    % \scriptsize
    % \setlength{\tabcolsep}{6pt}
    \setlength{\tabcolsep}{15pt}
    % \renewcommand{\arraystretch}{1.2}
    % \resizebox{0.9\textwidth}{!}{%
    % \begin{tabular}{@{}lcccc@{}}
    \begin{tabularx}{0.9\textwidth}{@{}lcccc@{}}
    \toprule
    \multicolumn{1}{c}{\makecell[cc]{Method}} & \makecell[cc]{BEDLAM2\\MPJPE / PA / Cam.} & \makecell[cc]{CMU Panoptic\\MPJPE / PA / Cam.} & \makecell[cc]{HuMMan\\MPJPE / PA / Cam.} & \makecell[cc]{EgoBody\\MPJPE / PA / Cam.} \\
    \midrule
    3DB & 63.8 / 38.0 / 266.7 & 60.9 / 33.6 / 122.9 & 41.2 / 31.0 / 76.7 & 57.5 / 32.1 / 97.4 \\
    3DB + depth correction & 63.8 / 38.0 / 184.5 & 60.9 / 33.6 / 102.0 & 41.2 / 31.0 / 52.7 & 57.5 / 32.1 / 70.2 \\
    \midrule
    Ours (RGB) & 50.6 / 31.2 / 188.2 & 47.1 / 28.2 / 104.6 & 43.6 / 31.4 / 81.2 & 49.8 / 30.4 / 90.1 \\
    Ours (RGB-Ground) & \textbf{49.4} / 31.2 / \textbf{147.0} & \textbf{44.0} / 28.2 / \textbf{78.5} & 40.2 / 31.4 / 50.1 & 52.5 / 30.4 / 96.0 \\
    Ours (RGB-D) & 50.4 / \textbf{30.2} / 166.1 & 47.7 / \textbf{28.0} / 98.5 & \textbf{37.9} / \textbf{30.4} / \textbf{39.7} & \textbf{46.8} / \textbf{29.7} / \textbf{47.5} \\

    \bottomrule 
    % \end{tabular}%
    \end{tabularx}%
    % }
    \par\vspace{5pt}
    \begin{minipage}{0.9\textwidth}
    \footnotesize
    \textit{Notes.} Entries are MPJPE / PA-MPJPE / camera-local MPJPE in mm. All methods use identical oracle person boxes and ground-truth intrinsics on held-out test data. Depth correction denotes the optional training-free pelvis-distance correction with depth data.
    \vspace{-10pt}
    \end{minipage}
\end{table*}

\subsection{Multi-Dataset Results}
\label{sec:benchmarks}
We evaluate on two benchmark groups. For 3DPW, EMDB, and RICH, depth maps are rendered from dataset-provided body meshes; for BEDLAM2, CMU Panoptic, HuMMan, and EgoBody, we use aligned depth from each dataset pipeline. Table~\ref{tab:standard} uses GT or MoGe-2-estimated intrinsics to show sensitivity to calibration. Table~\ref{tab:rgbd} compares the input modes under matched samples and boxes.

Table~\ref{tab:standard} separates the effects of RGB fine-tuning, depth input, and camera calibration. Compared with the 3DB model, our RGB mode provides an overall improvement, particularly in PA-MPJPE and camera-local MPJPE. The magnitude varies across datasets because the three metrics retain different components of scale, orientation, and placement error.

Adding mesh-rendered depth produces the clearest and most consistent improvement across all three metrics. The gains in MPJPE and PA-MPJPE indicate better body-joint reconstruction, while the substantially larger gains in camera-local MPJPE show that depth is especially effective for resolving physical scale and camera-relative placement. The benefit is retained with MoGe-2-estimated intrinsics, although GT intrinsics generally provide better camera-local accuracy, indicating that calibration remains important. Because this table uses mesh-rendered depth, it isolates the value of aligned depth under controlled conditions; results using dataset-provided or registered depth are reported in Table~\ref{tab:rgbd}.

In Table~\ref{tab:rgbd}, comparing released 3DB with our RGB mode isolates the effect of fine-tuning, while comparisons with RGB-D and RGB-Ground evaluate the contributions of measured depth and ground geometry. Fine-tuning improves all three metrics on BEDLAM2, CMU Panoptic, and EgoBody, but slightly degrades them on HuMMan. Relative to RGB, RGB-D reduces camera-local MPJPE by 11.7\%, 5.8\%, 51.1\%, and 47.3\% across the four datasets. PA-MPJPE improves by only 0.7--3.2\%, indicating that depth primarily improves metric scale and camera-relative placement while providing smaller gains in body-joint reconstruction.

Ground geometry reduces camera-local MPJPE by 21.9\%, 25.0\%, and 38.3\% on BEDLAM2, CMU Panoptic, and HuMMan, respectively, but increases it by 6.5\% on EgoBody. Its unchanged PA-MPJPE indicates that these gains arise mainly from metric scale and placement rather than articulated pose. The training-free depth correction similarly reduces camera-local error by 17.0--31.3\% while leaving both pelvis-relative metrics unchanged, as expected for a camera-distance adjustment. RGB-Ground performs best on BEDLAM2 and CMU Panoptic, whereas RGB-D performs best on HuMMan and EgoBody, showing that the two metric cues are complementary but that ground-based correction depends on dataset and contact conditions.

\begin{figure*}[t!]
\centering
\includegraphics[width=0.9\textwidth]{
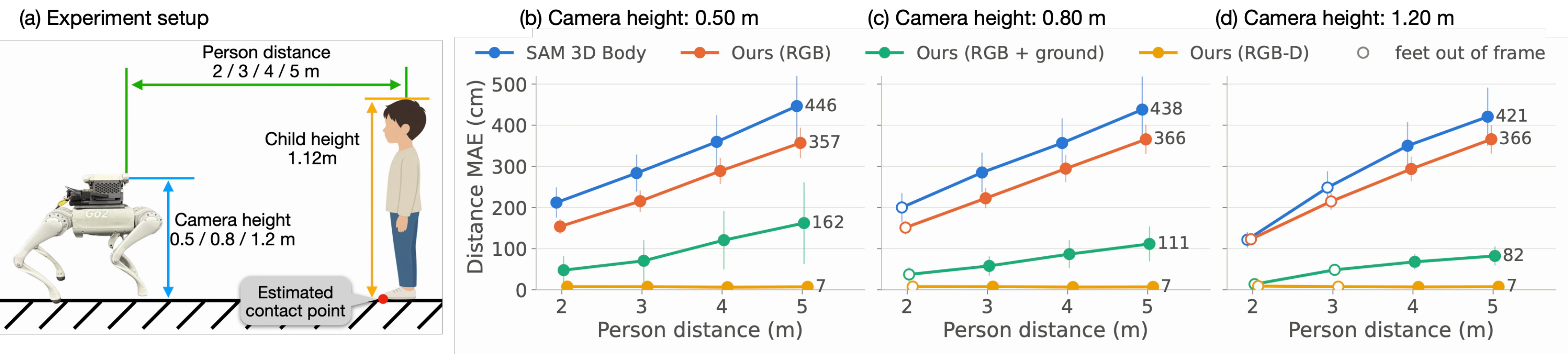
}
\vspace{-5pt}
\caption{Isaac Sim distance evaluation for a \SI{1.12}{\meter}-tall avatar. (a) Experiment setup. (b)--(d) Distance MAE at camera heights of \SIlist{0.5;0.8;1.2}{\meter}.}
\vspace{-10pt}
\label{fig:robot-experiment}
\end{figure*}

\subsection{CMU Panoptic Toddler Results}
\label{sec:annyfit}

We compare against Anny-Fit~\cite{annyfit}, an all-age fitting method that uses VLM-estimated age and predicted metric depth, on two CMU Panoptic toddler sequences viewed from two camera heights. Neither sequence is used for training. Anny-Fit lowers toddler MPJPE relative to 3DB and our RGB mode but raises PA-MPJPE, and it doubles the camera-local error of 3DB on the adults in the same scenes (Table~\ref{tab:annyfit}). Without age estimation or per-image optimization, our ground and depth modes reduce toddler camera-local MPJPE from \SI{2001}{} with 3DB to \SI{451}{} and \SI{97}{\milli\meter}; the latter is 91\% below Anny-Fit, and RGB-D is also best on adults.

\begin{table}[!t]
\caption{Results on CMU Panoptic toddler sequences.}
\label{tab:annyfit}
\centering
\scriptsize
\setlength{\tabcolsep}{18pt}
% \resizebox{\columnwidth}{!}{%
\begin{tabularx}{\columnwidth}{@{}lcc@{}}
% \begin{tabular}{@{}lcc@{}}
\toprule
\multicolumn{1}{c}{\makecell[cc]{Method}} & \makecell[cc]{Toddler\\MPJPE / PA / Cam.} & \makecell[cc]{Adult\\MPJPE / PA / Cam.} \\
\midrule
3DB & 218 / \textbf{35} / 2001 & 58 / 29 / 108 \\
Anny-Fit & 117 / 41 / 1093 & 82 / 41 / 219 \\
\midrule
Ours (RGB) & 196 / 36 / 2027 & \textbf{41} / 26 / 58 \\
Ours (RGB-Ground) & 78 / 36 / 451 & 43 / 26 / 74 \\
Ours (RGB-D) & \textbf{60} / 36 / \textbf{97} & \textbf{41} / \textbf{25} / \textbf{49} \\
\bottomrule
% \end{tabular}%
\end{tabularx}%
% }
\vspace{-10pt}
\end{table}

\subsection{OpenCap Results}
\label{sec:additional-comparison}

\textbf{Protocol.} OpenCap~\cite{opencap} estimates kinematics from multiple calibrated smartphone views using a subject-scaled OpenSim model. We evaluate on its LabValidation dataset, which provides five iPhone views and optical motion-capture ground truth for ten adults. Camera-local joint centers are obtained by forward kinematics of the motion-capture inverse-kinematics solution. Because the released extrinsics and video timing are imperfectly aligned, we estimate a per-session rigid correction and per-clip temporal offset using a keypoint detector unrelated to the compared methods.

All methods are evaluated on the same 12 limb joints, using the hip midpoint as the pelvis and MHR70 acromion landmarks for the shoulders. Excluding drop-jump trials and one subject without video leaves 90 trials (33.5k frames) from nine subjects; errors are averaged within trials and then across trials. OpenCap uses its released two-camera HRNet estimates. 3DB and our method use one calibrated $70^\circ$ view with oracle boxes and no subject-specific information. The RGB-Ground mode additionally receives the ground plane obtained from the corrected camera calibration and gravity direction.

\textbf{Results.} Table~\ref{tab:additional-comparison} shows that RGB
fine-tuning improves all three metrics over 3DB, reducing MPJPE, PA-MPJPE, and camera-local MPJPE by \SIlist{2.5;2.3;4.2}{\milli\meter}, respectively. The camera-local gain indicates that learned adaptation also improves global scale and placement without metric cues. Ground geometry provides a larger complementary gain, reducing camera-local MPJPE from \SI{133.9}{} to \SI{96.3}{\milli\meter} (28.1\%) and MPJPE from \SI{46.3}{} to \SI{42.8}{\milli\meter} while leaving PA-MPJPE essentially unchanged. This pattern shows that the ground cue primarily improves physical scale and placement rather than aligned pose. Despite using one view and no subject-specific model, RGB + Ground comes within \SI{3.9}{\milli\meter} MPJPE of two-view OpenCap and closes approximately 40\% of the camera-local gap between RGB and OpenCap. OpenCap's remaining advantage is expected from its two calibrated views and subject-specific musculoskeletal scaling.

\begin{table}[!t]
\caption{Comparison on the OpenCap dataset (mm; $\downarrow$ lower is better).}
\label{tab:additional-comparison}
\centering
\scriptsize
\setlength{\tabcolsep}{4pt}
\resizebox{\columnwidth}{!}{%
\begin{tabular}{@{}llccc@{}}
\toprule
\multicolumn{1}{c}{Method} & \multicolumn{1}{c}{Views} & \multicolumn{1}{c}{MPJPE} & \multicolumn{1}{c}{PA-MPJPE} & \multicolumn{1}{c}{Cam.-local MPJPE} \\
\midrule
OpenCap & Multi & \textbf{38.9} & \textbf{30.7} & \textbf{41.1} \\
3DB & Single & 48.8 & 34.6 & 138.1 \\
Ours (RGB) & Single & 46.3 & 32.3 & 133.9 \\
Ours (RGB-Ground) & Single & 42.8 & 32.4 & 96.3 \\
\bottomrule
\end{tabular}%
}
\vspace{-10pt}
\end{table}

\subsection{Camera-to-Person Distance Estimation}
\label{sec:robot-distance}

\textbf{Protocol.} To evaluate distance errors caused by monocular body-scale ambiguity, we use a \SI{1.12}{\meter}-tall small-stature avatar performing multiple poses in Isaac Sim. We vary the horizontal camera-to-person distance over \SIlist{2;3;4;5}{\meter} and the camera height over \SIlist{0.5;0.8;1.2}{\meter}. We report the MAE of the estimated horizontal camera-to-person distance for SAM 3D Body and our RGB, RGB-Ground, and RGB-D modes. The controlled simulation provides exact depth, camera calibration, and ground geometry.

\textbf{Results.} Fig.~\ref{fig:robot-experiment} shows that monocular camera-to-person distance errors increase approximately linearly with distance. At \SI{5}{\meter}, SAM 3D Body reaches \SIrange{421}{446}{\centi\meter} MAE, while RGB fine-tuning reduces the error to \SIrange{357}{366}{\centi\meter} but retains the same distance-dependent trend. Ground geometry provides a substantially stronger constraint, reducing the \SI{5}{\meter} error to \SIlist{162;111;82}{\centi\meter} at camera heights of \SIlist{0.5;0.8;1.2}{\meter}, respectively. Its improvement with camera height is consistent with the greater geometric stability of a higher camera-ground baseline. RGB-D maintains approximately \SI{7}{\centi\meter} MAE across all tested distances and camera heights, including configurations in which the feet leave the image. These results show that RGB fine-tuning reduces but does not eliminate distance errors caused by body-scale ambiguity, ground geometry provides an intermediate constraint, and depth gives the most stable distance estimates.

\subsection{Ablation}
\label{sec:ablations}

We perform one-at-a-time inference-time pathway knockouts on HuMMan with sensor depth and EMDB with mesh-rendered depth, keeping all other components fixed and disabling the optional pelvis-distance correction in every condition. Table~\ref{tab:depth-ablation} reports $\Delta=E_{\mathrm{disabled}}-E_{\mathrm{full}}$; positive values indicate degradation after removing a pathway. This analysis measures the jointly trained model's reliance on each pathway, not the effect of retraining without it.

\begin{table}[t]
\caption{Test-time depth-pathway knockouts. Entries are changes in MPJPE /
PA-MPJPE / camera-local MPJPE after removal (mm); positive values indicate
that the pathway is beneficial.}
\label{tab:depth-ablation}
\centering
\scriptsize
\setlength{\tabcolsep}{3pt}
\resizebox{\columnwidth}{!}{%
\begin{tabular}{@{}lcc@{}}
\toprule
\multicolumn{1}{c}{\makecell[cc]{Disabled module}} & \makecell[cc]{HuMMan\\$\Delta$MPJPE / PA / Local} & \makecell[cc]{EMDB\\$\Delta$MPJPE / PA / Local} \\
\midrule
Joint refiner
    & +2.10 / +0.62 / +1.55
    & +3.22 / +1.10 / +2.41 \\
Backbone fusion
    & +0.52 / +0.25 / +0.43
    & +1.71 / +0.95 / +1.37 \\
Depth-surface correction
    & +2.87 / 0.00 / +40.99
    & +3.90 / 0.00 / +49.46 \\
\bottomrule
\end{tabular}%
}
\vspace{-10pt}

\end{table}

The depth-surface correction accounts for most of the camera-local placement gain: disabling it increases camera-local MPJPE by \SI{41.0}{\milli\meter} on HuMMan and \SI{49.5}{\milli\meter} on EMDB, while PA-MPJPE is unchanged at the reported precision. Among the learned pathways, the joint refiner provides the largest pose improvement, followed by backbone fusion. These results indicate that surface correction primarily improves metric placement, whereas the refiner and backbone fusion improve joint configuration.

\section{Conclusion}

We presented a unified extension of 3DB for metric 3D human pose estimation with optional depth and ground geometry. A shared model supports RGB, RGB-Ground, and RGB-D inference, combining learned RGB-D feature fusion and depth-guided pose refinement with deterministic metric corrections. Across the benchmarks, the addition of depth consistently reduced camera-local MPJPE, while its smaller effect on pelvis-relative and Procrustes-aligned errors showed that the principal gain is in physical scale and camera-relative placement. Ground geometry provided substantial improvements when its support-plane assumptions held, but remained dataset-dependent. The controlled small-stature experiment further exposed the monocular size--distance ambiguity and showed that adding depth largely removed its distance-dependent error.

\textbf{Limitations.} Performance remains sensitive to missing or misregistered depth and calibration error. Ground correction assumes known gravity, camera height, planar support, and valid contact, while the deterministic corrections use calibrated thresholds. The depth branch also increases computation. 

\textbf{Future work.} We will investigate uncertainty-aware fusion, calibration robustness, temporal or multi-view cues, and more efficient depth processing.

\section*{ACKNOWLEDGMENT}

This material is based upon work supported by the National Science Foundation (NSF) under Grant Nos. 2427788, 2541914, 2529277, and 2442120; the Electronics and Telecommunications Research Institute (ETRI); and the NVIDIA Academic Grant Program. This work was also supported by a Pilot Grant from the Restore Center, which is funded by the National Institutes of Health (NIH) under Grant No. P2CHD101913.

\bibliographystyle{IEEEtran}
\bibliography{bib}

\end{document}